**Automatic Defect Detection of Print Fabric Using Convolutional Neural Network**

Samit Chakraborty[1*], Marguerite Moore[1], Lisa Parrillo-Chapman[1]

[1]Wislon College of Textiles, North Carolina State University, Raleigh, NC, USA

(*Corresponding author: schakr22@ncsu.edu)

**Abstract**

Automatic defect detection is a challenging task because of the variability in texture and type of fabric defects. An effective defect detection system enables manufacturers to improve the quality of processes and products. Automation across the textile manufacturing systems would reduce fabric wastage and increase profitability by saving cost and resources. There are different contemporary research on automatic defect detection systems using image processing and machine learning techniques. These techniques differ from each other based on the manufacturing processes and defect types. Researchers have also been able to establish real-time defect detection system during weaving. Although, there has been research on patterned fabric defect detection, these defects are related to weaving faults such as holes, and warp and weft defects. But, there has not been any research that is designed to detect defects that arise during such as spot and print mismatch. This research has fulfilled this gap by developing a print fabric database and implementing deep convolutional neural network (CNN).

## 1. Introduction

Textile and apparel industries primarily initiate quality inspection of raw materials and fabrics through human visual inspection. Here any deviation of textile properties from a standard is classified as defect. Quality inspectors randomly examine fabric rolls visually after these rolls are loaded onto inspection machines without the aid of technology [1]. However, reliance on the manual labor based inspection cause different problems in quality control because of human fatigue and potential inattentiveness [2]. According to Essid et al. manufacturing industries have now started adopting computer vision (CV) to implement automatic product inspection using machine learning and image processing algorithms at multiple stages of the manufacturing process [3]. Materials for garment production typically account for the largest proportion of production cost and are therefore critical to profitable production [2], [4], [5]. Moreover, real-time defect detection executed through autonomous devices can overcome limitations inherent to traditional human inspection systems through automatic detection of variations from a pre-defined visual standard [2] [3].

Manufacturing processes are never free of defects. A human being can detect only 60 to 70 percentages of total textile fabric defects [1]. However, the CV based automatic defect detection system (ADD) can detect 90 percentages of textile and apparel defects [6]. Textile defects and faulty manufacturing processes cause various problems in the production line as well as low quality fabric and clothing [7]. Therefore, higher it has now become mandatory to implement CV based ADD textile manufacturing industry. Nevertheless, the application of real-time ADD has not been widespread across the textile manufacturing supply chain system. For instance, there has been no printed fabric dataset that is developed based on original fabrics collected form the industry as like as Tilda [8]. Moreover, printed fabric manufacturing is one of the costly processes and its waste can cause to a high loss. Therefore, this research focused on developing a deep convolutional neural network (DCNN) based automatic defect detection system for printed fabric.

Therefore, the purpose of this research is to-

1. Develop a high-quality print fabric database, which covers both defect free and defective knit fabric images.
2. Detect print spot and print mismatch defects in the fabric automatically using Convolutional Neural Network

## 2. Automatic Fabric Defect Detection with Deep Convolutional Neural Network (DCNN)

### 2.1. The Proposed Deep Convolutional Neural Network Architecture

In recent years, different types of deep learning based machine learning models have been developed, such as deep-stacked auto-encoder (DAE), deep convolutional neural network (DCNN) and deep belief network (DBN) Among these models, CNN has been more popular and successful [9]–[11]. CNN consists of neurons or nodes with weights (w) and biases (b) [12], [13]. CNN comprises different layers named as- input, output and hidden layers. The hidden layers usually comprises of a multiple convolutional, pooling and fully connected layers [14].

*Convolution Layer*

In the convolution layer, a filter is passed over the input image is convoluted by the filters to form a feature map and predict the class or label to which a feature belong [10], [15]. Here in Figure 2, the matrix in the furthest left is a numerical representation of a random image and the matrix in the middle represents a convolutional matrix or convolutional filter. The pixel values covered by the original image are multiplied with corresponding filter or convolutional matrix values and their products are summed to get a specific value in the feature map or convoluted image as shown in the rightmost corner of Figure 1 [16].

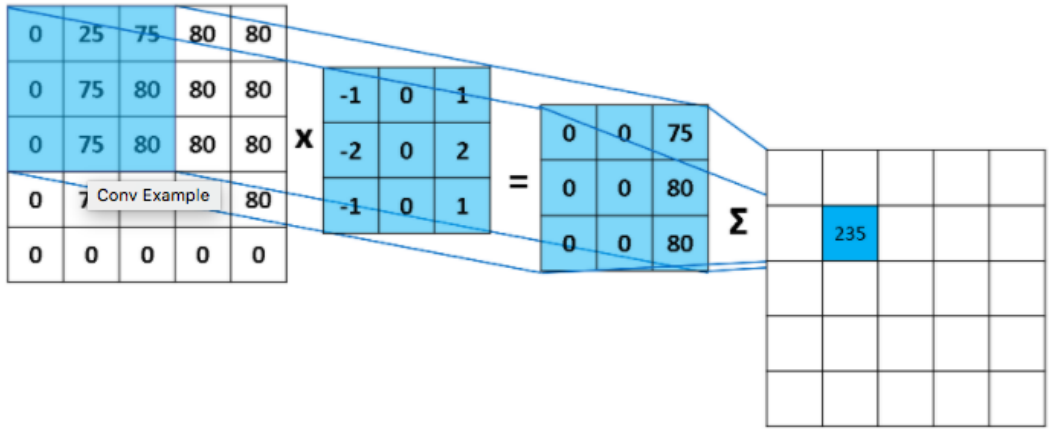

Figure 1: A convolutional step [16]

A convolution layer (shown in Figure 2) can be expressed through equation 1, where $x_i$ is the $i^{th}$ input, weight$_i$ is the weight on $i^{th}$ input, bias on $i^{th}$ input and y is the output for $i^{th}$ input. Here '*' indicates convolution operation and '$f()$' indicates activation function [10].

$$y = f(\sum_i x_i * weight_i + bias) \ \dots\dots\dots\dots\dots\dots\dots\dots\dots\dots\dots\dots\dots\dots\dots\dots\dots(1)$$

*Pooling Layer*

Pooling layer in CNN architecture follows convolutional layer. It reduces the size of the feature map [10], [15], which subsequently decreases the amount of information or feature parameters [13]. Meantime, it also controls the important information extracted from convoluted image [10], [15]. There can be multiple sets of convolution and pooling layers in a DCNN based on the convolution of images (Figure 2).

*Fully connected layer*

Fully connected layer (FCL) receives the output from the pooling layer and implements them by categorizing the images with different labels and flattens them into probability values representing a class [10], [13], [15].

*Output layer*

The output layer presents the final probability values or scores, which help to classify the images (Figure 2) [10], [13], [15]

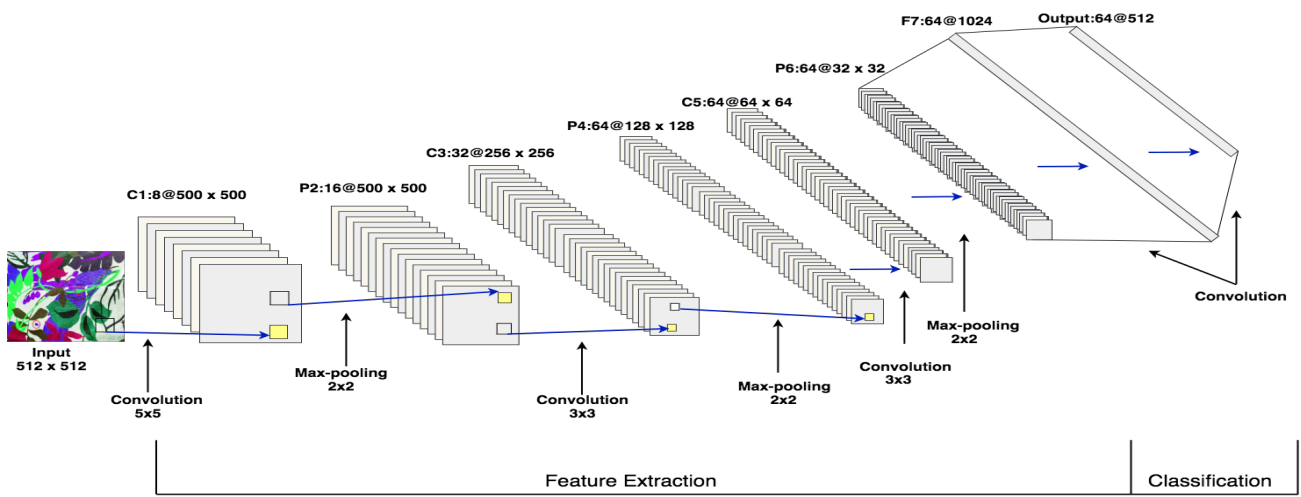

**Figure 2: A CNN architecture representing various components and operations**

## 3. Research Methodology and Experimental Analysis

### 3.1. Data Collection and Equipment Setup

Three hundred and eighty fabric images were collected from industry members. Initially, the total dataset was split into 40 (train): 30 (validation): 30 (test) ratio and then image augmentation (5x) was carried out to expand the train dataset from 152 to 760. Images were labelled to annotate the defect free fabric image as 0, color spot as 1 and misprint as 2. CNN algorithm was used and the parameters were hypertuned to run the model using python code. After tuning the CNN, we have used the same set of hyperparameters and dense layers with different network architectures including VGG16 and VGG19.The proposed CNN model was compared with VGG16 and VGG19 models. While running all these experiments, we had to use a graphics-processing unit (GPU), which supports faster performance with the TensorFlow-GPU package. Therefore, we used an NVIDIA Quadro GV100 GPU containing 12 GB HBM2 memory with a dual 10-core 2.1 GHz Intel processor and 64GB 2600MHz DDR4 ECC memory to speed up the model training.

## 3.2. Results and Key Findings

At first, the learning rate was adjusted for different combinations of hyperparameters such as number of hidden layers, dropout probability, regularization and activation function. The learning rate yielding the best validation accuracy was chosen. Then, batch size was adjusted for multiple combinations of rest of the hyperparameters keeping constant the best learning rate. The similar process was iterated while adjusting the rest of the hyperparameters. A simple iteration (10 epochs each) was utilized on individual parameters sequentially to search for a value that gives the highest validation accuracy. After training our custom CNN (training and validation loss and accuracy are shown in Fig. 3), the test dataset was used to generate the class predictions. Fig. 3 shows the training and validation loss for sigmoid activation function.

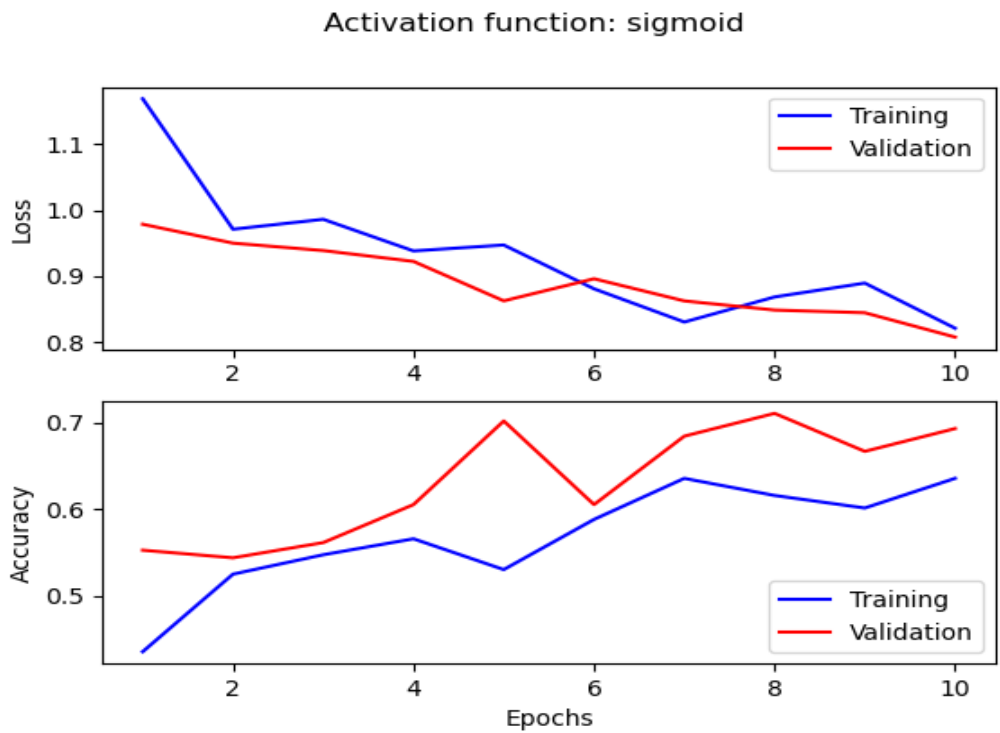

Fig. 3. Plots of loss and accuracy vs. epochs during the training of the best model (VGG16)

To give a comprehensive display of the network performance, we have checked the commonly used metrics: accuracy (2) and geometric mean of precision (3) and recall (4), which are respectively defined as the proportion of correct predictions. The confusion matrix generated from the custom CNN with the best activation function has been shown in Fig. 4.

$$Accuracy = \frac{TP+TN}{TP+TN+FP+FN} \quad \dots\dots\dots\dots\dots\dots\dots\dots\dots\dots\dots\dots\dots\dots\dots\dots(2)$$

$$Precision = \frac{TP}{TP+FP} \quad \dots\dots\dots\dots\dots\dots\dots\dots\dots\dots\dots\dots\dots\dots\dots\dots\dots\dots(3)$$

$$Recall = \frac{TP}{TP+FN} \quad \dots\dots\dots\dots\dots..\dots\dots\dots\dots\dots\dots\dots\dots\dots\dots\dots\dots(4)$$

where, TP = True Positive, TN = True Negative, FP = False Positive, FN = False Negative.

Fig. 4 shows the confusion matrix of VGG16, where the model showed 56%, 73% and 62% accuracy on detecting defect free, color spot and misprint images respectively. The VGG16, VGG19 and proposed DCNN achieved test accuracy equivalent to 62.28%, 60% and 48.5% respectively. Since, this dataset contains very complex images and has never been used before, there was no benchmark to compare the performances.

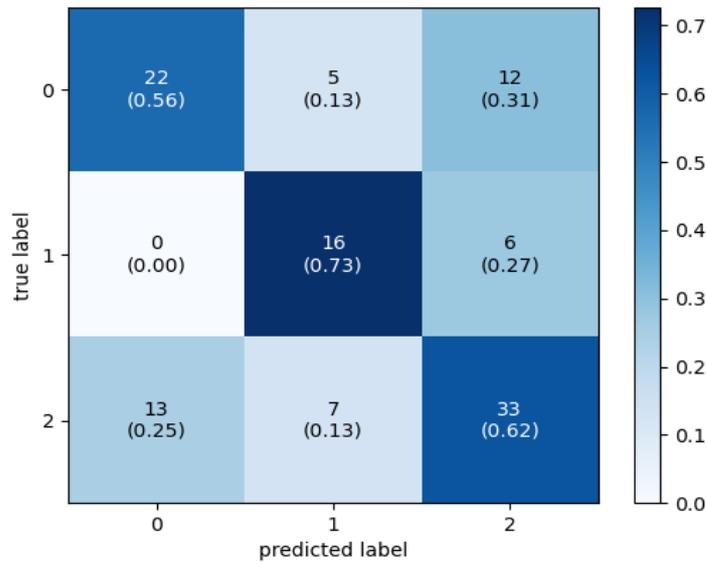

Fig.4: Confusion Matrix of predictions on the testing dataset with the best model (VGG16 Sigmoid)

## 6. Conclusion

This research project was a good opportunity to explore the various pre-trained networks in the TensorFlow Keras Library. The model achieved a high score considering the type and complexity, image pre-processing, data augmentation and rejection of fabric images. Testing of this model during in a fully operational production setting in order for simulate real-time defect detection would be a first ever approach in the textile printing industry. In future studies a larger dataset could be used and more defect types could be included.

**Reference**

[1] W. K. Wong and J. L. Jiang, "Computer vision techniques for detecting fabric defects," in *Applications of Computer Vision in Fashion and Textiles*, Elsevier, 2018, pp. 47–60.

[2] A. Kumar, "Computer-Vision-Based Fabric Defect Detection: A Survey," *IEEE Trans. Ind. Electron.*, vol. 55, no. 1, pp. 348–363, Jan. 2008, doi: 10.1109/TIE.1930.896476.

[3] O. Essid, H. Laga, and C. Samir, "Automatic detection and classification of manufacturing defects in metal boxes using deep neural networks," *PLoS ONE*, vol. 13, no. 11, p. e0203192, Nov. 2018, doi: 10.1371/journal.pone.0203192.

[4] K. Hanbay, M. F. Talu, and Ö. F. Özgüven, "Fabric defect detection systems and methods—A systematic literature review," *Optik*, vol. 127, no. 24, pp. 11960–11973, 2016, doi: 10.1016/j.ijleo.2016.09.110.

[5] H. Y. T. Ngan, G. K. H. Pang, and N. H. C. Yung, "Automated fabric defect detection—A review," *Image and Vision Computing*, vol. 29, no. 7, pp. 442–458, Jun. 2011, doi: 10.1016/j.imavis.2011.02.002.

[6] R. Furferi and L. Governi, "Machine vision tool for real-time detection of defects on textile raw fabrics," *Journal of the Textile Institute*, vol. 99, no. 1, pp. 57–66, Jan. 2008, doi: 10.1080/00405000701556426.

[7] M. Eldessouki, "Computer vision and its application in detecting fabric defects," in *Applications of Computer Vision in Fashion and Textiles*, Elsevier, 2018, pp. 61–101.

[8] Tilda, "Datasets," *lmb.informatic*, 1996. https://lmb.informatik.uni-freiburg.de/resources/datasets/tilda.en.html (accessed Nov. 09, 2020).

[9] N. T. H. Anh and B. C. Giao, "An Empirical Study on Fabric Defect Classification Using Deep Network Models," in *Future Data and Security Engineering*, vol. 11814, T. K. Dang, J. Küng, M. Takizawa, and S. H. Bui, Eds. Cham: Springer International Publishing, 2019, pp. 739–746.

[10] C. Gao, J. Zhou, W. K. Wong, and T. Gao, "Woven Fabric Defect Detection Based on Convolutional Neural Network for Binary Classification," in *Artificial Intelligence on Fashion and Textiles*, vol. 849, W. K. Wong, Ed. Cham: Springer International Publishing, 2019, pp. 307–313.

[11] W. Ouyang, B. Xu, J. Hou, and X. Yuan, "Fabric Defect Detection Using Activation Layer Embedded Convolutional Neural Network," *IEEE Access*, vol. 7, pp. 70130–70140, 2019, doi: 10.1109/ACCESS.2019.2913620.

[12]    K. Hanbay, M. F. Talu, Ö. F. Özgüven, and D. Öztürk, "REAL-TIME DETECTION OF KNITTING FABRIC DEFECTS USING SHEARLET TRANSFORM," *Tekstil ve Konfeksiyon*, vol. 29, no. 1, Art. no. 1, Mar. 2019, doi: 10.32710/tekstilvekonfeksiyon.482888.

[13]    Stanford, "CS231n Convolutional Neural Networks for Visual Recognition," *stanford.edu*, 2020. https://cs231n.github.io/convolutional-networks/) (accessed Oct. 31, 2020).

[14]    I. Goodfellow, Y. Bengio, and A. Courville, *Deep learning*. Cambridge, Massachusetts: The MIT Press, 2016.

[15]    Missinglink, "Convolutional Neural Networks Fully Connected Layers in Convolutional Neural Networks: The Complete Guide," *missinglink.ai*, 2020. https://missinglink.ai/guides/convolutional-neural-networks/fully-connected-layers-convolutional-neural-networks-complete-guide/.

[16]    Mlnotebook, "Convolutional Neural Networks - Basics An Introduction to CNNs and Deep Learning," *MLNOTEBOOK*, 2017. https://mlnotebook.github.io/post/CNN1/ (accessed Oct. 31, 2020).